# Syllable parsing in English and French*

Michael Hammond
University of Arizona
draft: May 25, 1995

## 0. Overview

In this paper I argue that Optimality Theory provides for an explanatory model of syllabic parsing in English and French. The argument is based on psycholinguistic facts that have been mysterious up to now. This argument is further buttressed by the computational implementation developed here. This model is important for several reasons. First, it provides a demonstration of how OT can be used in a performance domain.[1] Second, it suggests a new relationship between phonological theory and psycholinguistics.

The organization of this paper is as follows. First, I review the linguistic evidence for syllable structure in English. Then, experimental evidence is considered. The latter comes from tasks tapping into the units of speech perception by speakers of English and French, and constitutes an argument against the importation of rule-based syllabification into the psycholinguistic domain, suggesting that an alternative constraint-based approach might be more appropriate. We then review Optimality Theory (OT) and present an approach to syllabic parsing in terms of OT that can elegantly model the processing differences between English and French and account readily for the linguistic facts of English.[2] Finally, the model is implemented in a computer program using Perl. The implementation is interesting because i) it shows that the model proposed is computationally feasible, and ii) it demonstrates the use of several devices – serial constraint satisfaction and "local coding" – in reducing the candidate set.

## 1. Linguistic background

The classic argument for syllables in English comes from a consideration of basic distributional regularities concerning consonants and vowels. Specifically, it can be argued that the character of consonant clusters in the middle of English words is a function of the character of clusters at the edges of words. These regularities follow if it is assumed that words are exhaustively parsed into syllables.[3]

Consider, for example, the character of word-initial and word-final clusters in English containing the bilabial nasal [m].

---

* Thanks to Diane Ohala for much useful discussion.

[1] Dupoux, Hammond, Meador, & Ohala (1995) argue that OT can be modified to account for offline syllabification judgments.

[2] There are obviously linguistic facts of French at issue, but a full analysis of French syllabification would take us far afield. See the references cited below for discussion.

[3] In this paper, I do not treat the effect of vowel quality on syllabification. For a discussion of the role of vowel quality in the analysis of English syllabification, see Pulgram (1970). For discussion of these facts in OT terms, see Hammond (1993) and Dupoux, Hammond, Meador, & Ohala (1995).

(1)

| | | | |
|---|---|---|---|
| word-initial | N | m | *mint* |
| | N+C | my | *mule* |
| | C+N | sm | *small* |
| word-final | N | m | *come* |
| | N+C | mp | *limp* |
| | | mf | *triumph* |
| | | mz | *limbs* |
| | | md | *rhymed* |
| | | mt | *dreamt* |
| | C+N | rm | *harm* |
| | | lm | *helm* |
| | | ym | *time* |

All word-internal clusters containing [m] can be analyzed as instances of one of the above word-final clusters followed by some other licit word-initial cluster or one of the above word-initial clusters preceded by some other licit word-final cluster. Here are some examples.

(2)

| | word | analysis | onset as in |
|---|---|---|---|
| a | *hamster* | m + str | *string* |
| b | *bombard* | m + b | *book* |
| c | *camphor* | m + f | *fish* |
| | | mf + Ø | *apple* |
| d | *bumpkin* | mp + k | *kettle* |
| e | *complex* | mp + l | *lettuce* |
| | | m + pl | *please* |
| f | *membrane* | m + br | *brown* |
| g | *pamphlet* | m + fl | *flea* |
| h | *portmanteau* | rt + m | *heart* |
| i | *marshmallow* | rsh + m | *harsh* |
| j | *formless* | rm + l | *harm* |

There are no words in English of the following sort, where the cluster cannot be analyzed in this way.

(3)

| case | impossible analyses | | | |
|---|---|---|---|---|
| VmltV | *Vmlt+V | *Vml+tV | Vm+*ltV | V+*mltV |
| VsmsV | *Vsms+V | *Vsm+sV | Vs+*msV | V+*smsV |
| VhmV | *Vhm+V | *Vh+mV | V+*hmV | |

If we assume that all licit English words must be parsed into legal syllables, the absence of the clusters in (3) follows from the fact that there is no such parse.

(4) Exhaustive syllabification
Words must be exhaustively parsed into legal syllables.

Notice that this still leaves the parse indeterminate in many cases, e.g. (2c,e). In (5) are simpler examples involving only VCV sequences. In some of these examples, the syllabification is clear. For example, in (5a), the syllabification is clear because it is impossible to have words ending in [h]. (Since words cannot end in [h], it follows that syllables cannot end in [h] and that [VhV] sequences must be syllabified as [V.hV]. Some VCV sequences are indeterminate, however, because the relevant consonant can occur either at the beginning or end of the word, e.g. (5c).

(5) Determinacy of syllabification

a. unambiguous onset: *aha!* [a.ha]

b. unambiguous coda: *singing* [sɪŋ.ɪŋ]

c. indeterminate: *happy* [hæ.pi] or [hæp.i]

The same indeterminacy applies to larger sequences. The tables in (6) and (7) show two-consonant and three-consonant sequences where multiple syllabifications are possible. The different possible syllabifications are indicated with check marks in the appropriate rows.

(6)

| 2C | VstV | VltV | VtrV |
|---|---|---|---|
| VCC.V | √ | √ | * |
| VC.CV | √ | √ | √ |
| V.CCV | √ | * | √ |

(7)

| 3C | VstrV | VndwV | VrndV |
|---|---|---|---|
| VCCC.V | * | * | √ |
| VCC.CV | √ | √ | √ |
| VC.CCV | √ | √ | * |
| V.CCCV | √ | * | * |

There are, however, several sources of evidence for the affiliation of such consonants. One bit of evidence comes from nominal stress. Stress in English is quite complex, but one general fact is clear. Stress cannot occur on the third syllable from the right if the second syllable from the right is followed by two consonants where the first must close the syllable to the left.[4] Relevant data are presented in (8). The first column contains forms with antepenultimate stress where the penultimate vowel is followed by a single consonant. The second column contains forms with antepenultimate stress where the penultimate vowel is followed by two consonants that <u>can</u> be analyzed as beginning the final syllables. Finally, the third column illustrates forms with penultimate stress, where the penultimate vowel is followed by two consonants that <u>cannot</u> be analyzed as a licit syllable onset: antepenultimate stress is impossible with this last case.

[4]Forms with final syllables in *-y*, *-le*, *-re*, and various special affixes are exceptions to this generalization. See Chomsky & Halle (1968) for details.

(8)

| V.CV | V.CCV | VC.CV |
|---|---|---|
| Améri.ca | álge.bra | verán.da |
| cíne.ma | cócka.trice | agén.da |
| aspára.gus | ármi.stice | consén.sus |
| metrópo.lis | éndo.crine | synóp.sis |
| jáve.lin | sácri.stan | amál.gam |
| véni.son | órche.stra | utén.sil |

This pattern is readily accounted for in terms of syllables if the following assumptions are made. First, when consonants can be affiliated with either of two syllables, they are preferentially affiliated to the right.[5]

(9) Maximal Onset Principle [MOP]
In a $VC_0V$ sequence, affiliate as many consonants to the right as possible.

The MOP is limited by restrictions on syllable structure of the sort illustrated above. Second, heavy syllables – those closed by at least one consonant – cannot be skipped over by stress assignment.[6]

(10) Weight-to-Stress Principle [WSP]
Stress heavy syllables.

Words like *America*, *algebra*, and *veranda* are then syllabified as follows; heavy penults are underlined.

(11) A.me.ri.ca   al.ge.bra   ve.<u>ran</u>.da

Any other syllabification would violate the MOP or constraints on syllable structure.

There is, however, additional evidence that leads to an apparent paradox with respect to the MOP. This evidence comes from the distribution of aspiration in English. Voiceless stops in English can occur in at least two guises: aspirated ($C^h$) and unaspirated (C). Consider the distribution of aspiration in monosyllabic words. Word-initially such consonants are aspirated; after an [s] or word-finally, they are unaspirated.

(12)

| | | | |
|---|---|---|---|
| aspirated | $_{word}$[_V | *tack* | [tʰæk] |
| | $_{word}$[_C | *track* | [tʰræk] |
| unaspirated | _]$_{word}$ | *mat* | [mæt] |
| | $_{word}$[s_ | *stack* | [stæk] |

[5]This principle is originally due to Selkirk (1982).
[6]This affinity for stress is well-known. We adopt the formalization of Prince (1991) for currency.

The simplest generalization to account for these data is that voiceless stops are aspirated syllable-initially. Consider, however, the distribution of aspiration word-medially with respect to stress and compare this with the effect of stress in word-initial cases. Word-medially, aspiration occurs only if the following vowel is stressed; word-initially, stress is irrelevant.[7]

(13)

| | | | |
|---|---|---|---|
| word-medial | σ́_σ́ | *hotel* | [hòtʰɛ́l] |
| | σ́_σ̆ | *batter* | [bǽɾər] |
| | σ̆_σ́ | *attack* | [ətʰǽk] |
| | σ̆_σ̆ | *vanity* | [vǽnəɾi] |
| word-initial | word[_σ́ | *tacky* | [tʰǽki] |
| | word[_σ̆ | *tomorrow* | [tʰəmáro] |

The generalization that emerges is that voiceless stops are aspirated word-initially or medially before a stressed syllable. Kahn (1976) argues that this clumsy statement can be vastly simplified if it is assumed that there is stress-based resyllabification. Specifically, he proposes that a consonant before a stressless syllable affiliates to the left in defiance of the MOP.[8] If syllable structure is manipulated in this way, we can retain the simple characterization that aspiration occurs syllable-initially. The syllabifications required are given in (14) below.

(14)

| | | | |
|---|---|---|---|
| word-medial | σ́_σ́ | V.CV | [hòtʰɛ́l] |
| | σ́_σ̆ | VC.V | [bǽɾər] |
| | σ̆_σ́ | V.CV | [ətʰǽk] |
| | σ̆_σ̆ | VC.V | [vǽnəɾi] |
| word-initial | word[_σ́ | .CV | [tʰǽki] |
| | word[_σ̆ | .CV | [tʰəmáro] |

The paradox with respect to the MOP comes from the fact that the two syllabification schemes are contradictory. The stress facts require that VCV sequences syllabify as V.CV, but the aspiration facts require that VCv̆ sequences syllabify as VC.v̆ instead. Linguists have argued for several different derivational solutions to this paradox, but neither is compatible with the psycholinguistic evidence to be reviewed below.

One solution, due to Kahn (1976) and Selkirk (1983), is to i) syllabify in accord with the MOP, ii) assign stress, iii) resyllabify based on stress, iv) apply the aspiration rule.[9] This solution makes the prediction that at all points in the derivation after the first pass of syllabification, there should be an unequivocal syllabification either in accord with the MOP or not. The stress-based resyllabification rule can be given as (15).

---

[7]The voiceless alveolar stop [t] emerges as a flap [ɾ] in certain contexts when it is unaspirated. See Kahn (1976) for details.

[8]Kahn actually proposes that such consonants are affiliated both to the left and to the right, but we follow Selkirk (1983) in rejecting ambisyllabicity as a formal device here. See Meador & Ohala (1992) for another view though.

(15) Stress-Based Resyllabification
V.Cv̆ -> VC.v̆

A second solution to the paradox is to invoke higher order structures: metrical feet. In English, the metrical foot is comprised of a sequence of a stressed syllable followed by one or two stressless syllables (Hayes, 1981; Halle & Vergnaud, 1987; Hammond, 1991). Kiparsky (1979) argues that there is a laxing rule that applies foot-internally that precedes and bleeds the aspiration rule. Consonants in the environment for this laxing rule are underlined in (16).

(16)

| context | stress | foot boundaries | examples |
|---|---|---|---|
| word-medial | σ́_σ́ | V[CV | [hòtʰɛ́l] |
| | σ́_σ̆ | [VC̲V | [bǽɾəɾ] |
| | σ̆_σ́ | V[CV | [ətʰǽk] |
| | σ̆_σ̆ | [...VC̲V | [vǽnəɾi] |
| word-initial | $_{word}$[_σ́ | [CV | [tʰǽki] |
| | $_{word}$[_σ̆ | CV[ | [tʰəmáro] |

Flapping in words like *vanity* arises on the assumption that the metrical foot can extend to include the final syllable.

This latter approach to aspiration predicts that the syllable structure should be clear throughout the derivation. Specifically, MOP syllabification should hold at all levels.

## 2. Experimental background

In this section, we consider an experimental paradigm that sheds light on English syllabification. Cutler, Mehler, Norris, & Segui (1983, 1986) make use of a technique we can call *fragment monitoring* to show the role of syllable structure in speech processing. In this section, I show how both of the approaches above make incorrect predictions about the results of this task in English.

In the fragment monitoring task, subjects are presented a target string of sounds and asked to press a button as quickly as possible if that target begins a following word. For example, a French-speaking subject might be presented with the target *ba* or *bal* or *pa* and the word *balance*, and would be expected to press the button in the first two cases. Reaction times for positive responses are collected and analyzed.

[9]Kahn's and Selkirk's analyses differ in whether the resyllabification is complete or not. For Selkirk, resyllabification moves the consonant from the syllable to the right and into the syllable to the left. Kahn's analysis puts the consonant simultaneously in both syllables. This difference is not crucial for the issues discussed here and we assume Selkirk's position for representational convenience. See Meador & Ohala (1992) for experimental evidence bearing on this question.

Interestingly, there is an interaction between target type (CV vs. CVC) and word type (CV.CVX vs. CVC.CVX) in French. This is diagrammed below. Subjects respond more quickly to a CV target in a CV.CVX word and to a CVC target in a CVC.CVX word.

(17)

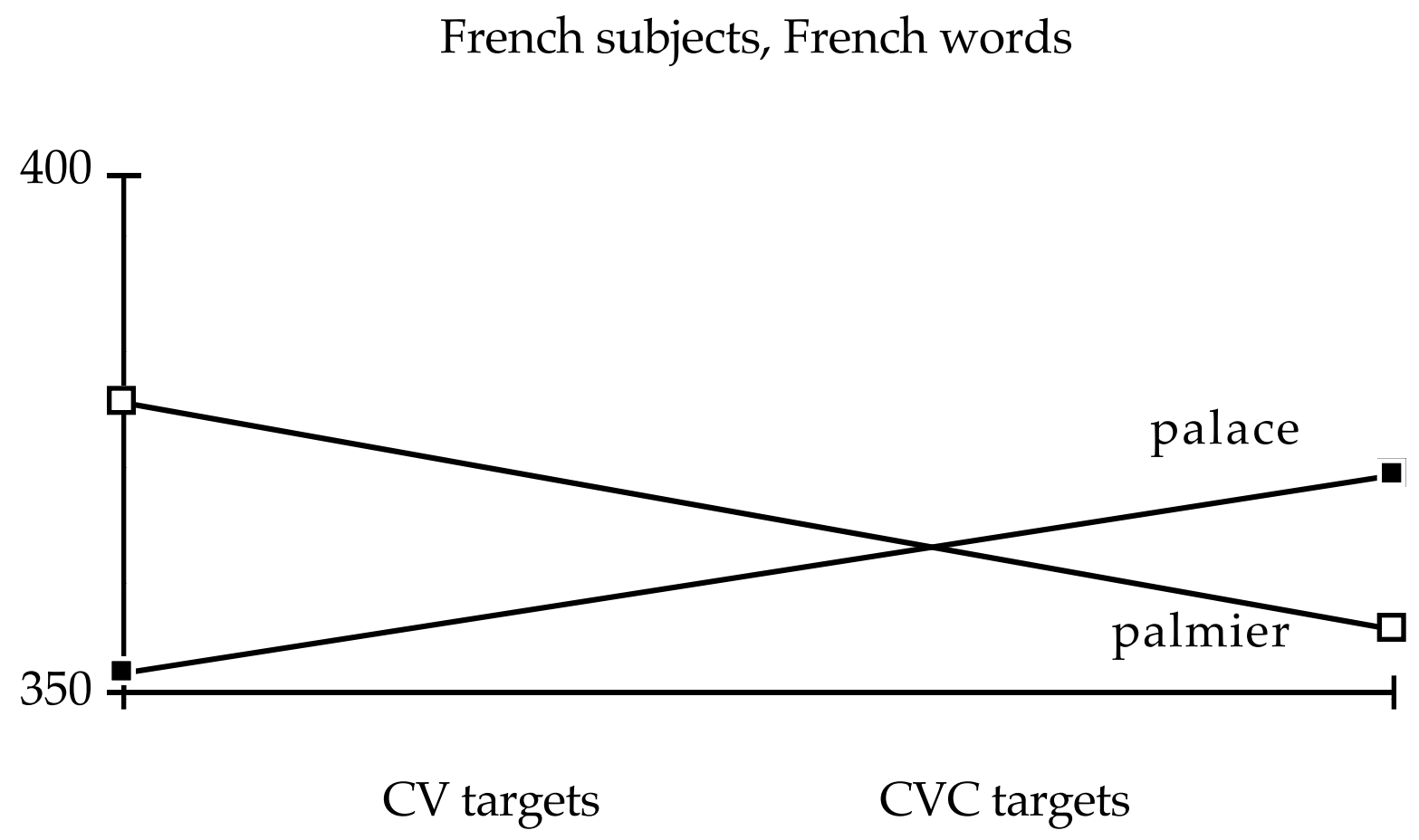

This difference suggests that syllable-sized units play a role in speech perception in French. The interaction shows that subjects are faster at identifying shared segmental content when there is a match in syllabic structure. This, in turn, implies that syllabic structure plays a role in the identification of words.

However, there is no such interaction when analogous materials are presented to English subjects. The table in (18) shows that there is no interaction between targets and carriers when English-speaking subjects are presented with targets like [bæ] and [bæl] and carriers like *balance* and *balcony*.

(18)

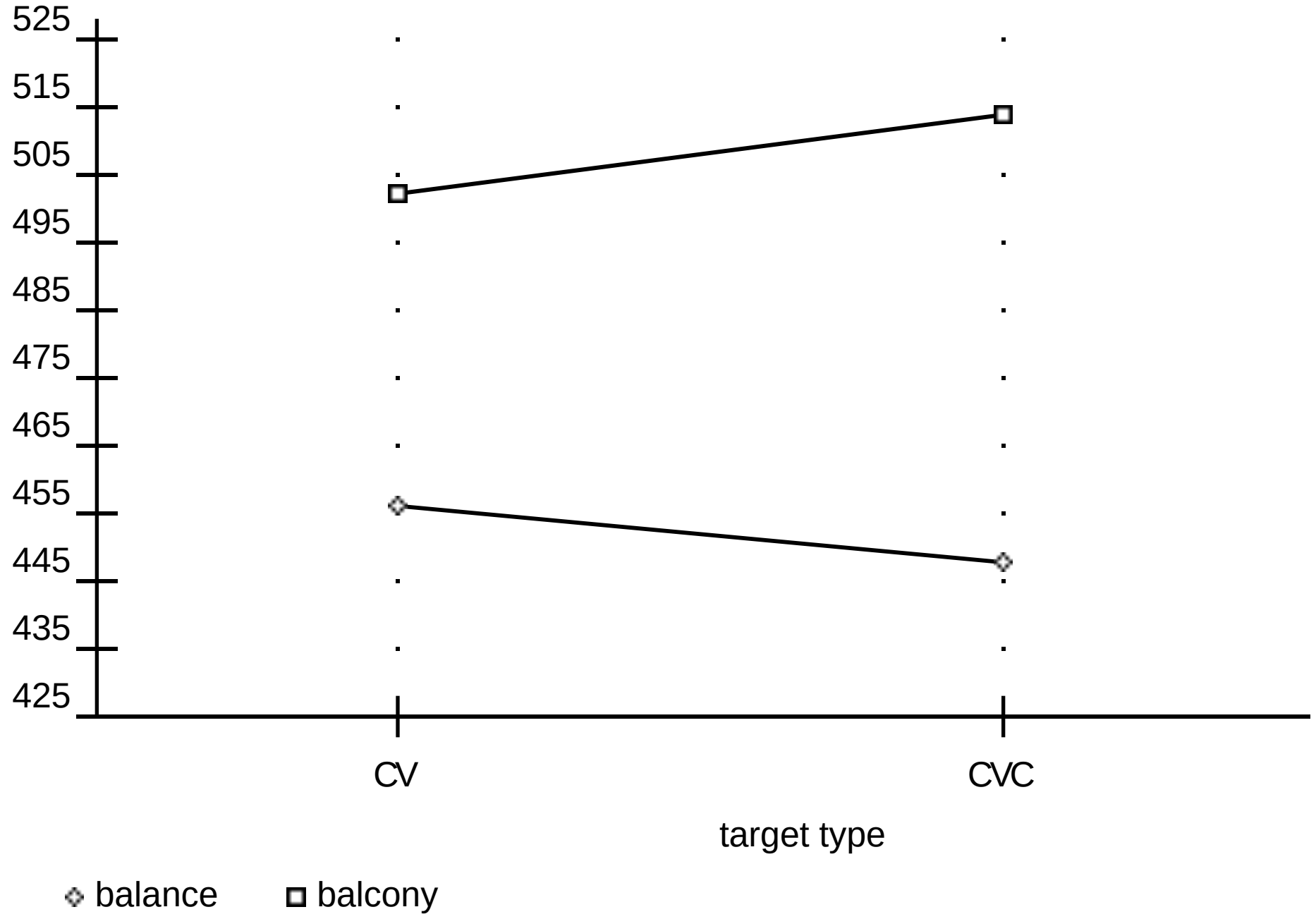

One might think that (18) establishes that there is no syllable used in the perception of English or that if the syllable is used in English, it is not the syllable predicted by linguistic theory. But it is still possible to maintain that the syllable used is that suggested by Kahn/Selkirk: the resyllabified syllable. On this view, we would expect no interaction with materials like *balance/balcony* because both first syllables should be closed since the following syllable is stressless: [bǽləns] vs. [bǽlkəni].

This, however, cannot be the case, Dupoux & Hammond (in sub) show that even where the Kahn/Selkirk resyllabified syllable makes different predictions, there is no interaction. Dupoux & Hammond contrast pairs like *climate/climax*, where the second syllables differ in whether they are stressed (leading to different affiliations because of the Kahn/Selkirk resyllabification rule), but there is still no effect: [kláymət] vs. [kláymæ̀ks].[10]

(19)

[10]The apparent small interaction in (19) is not significant and not replicated consistently in a series of experiments.

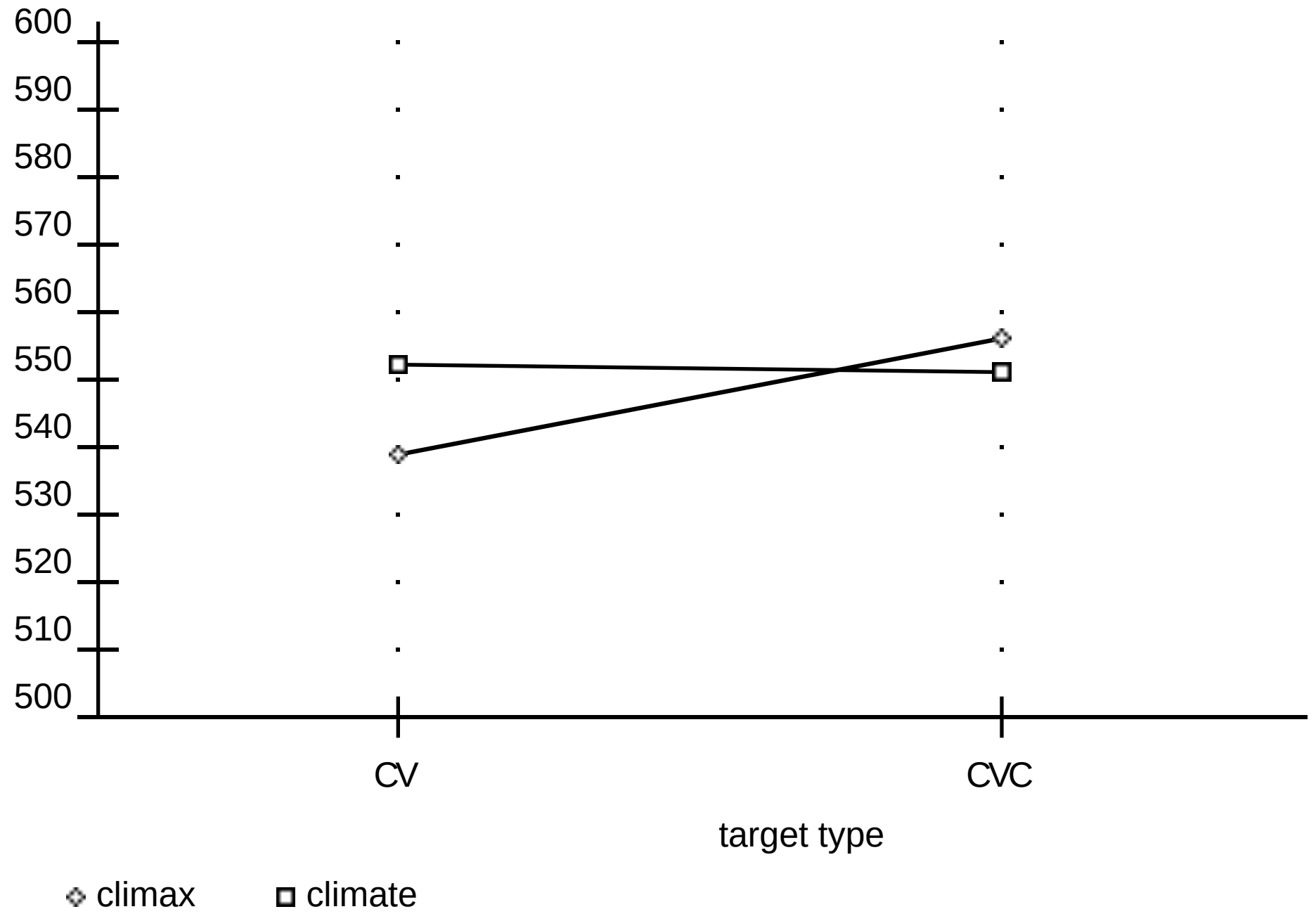

The results across these three experiments are summarized in (20).

| (20) | language | targets | carriers | interaction | reference |
|---|---|---|---|---|---|
| a | French | {pa / pal} | {pa.lace / pal.mier} | yes | Cutler, Mehler, Norris, & Segui (1983, 1986) |
| b | English | {bæ / bæl} | {bálănce / bálcŏny} | no! | Cutler, Mehler, Norris, & Segui (1983, 1986) |
| c | English | {klay / klaym} | {clímàx / clímăte} | no! | Dupoux & Hammond (in sub) |

There are two possible responses to this absence of an interaction with English subjects. First, syllables could be used in processing English, but they may be of a different sort from the syllables motivated on linguistic grounds. This would be certainly undesirable. All else being equal, we would hope that the syllables manipulated in processing to be the same as those motivated on linguistic grounds.

Another possibility would be that syllables are not used in speech perception in English at all, presumably because they are – in some objective sense – more complex than in French. This is what is suggested by Cutler et al.

I will propose a specific model of parsing syllables for English that i) accounts for the experimental effects and noneffects, ii) is fully in accord with the linguistic facts cited above, iii) puts some teeth to the Cutler et al. proposal above, and iv) is implemented computationally. The basic idea to be pursued is that the linguistic and psycholinguistic facts of English can be treated simultaneously if we adopt the view that syllabification is treated in terms of interacting constraints. This constraint-based syllabic parsing model that I propose is based on the emerging Optimality Theory (henceforth OT; Prince & Smolensky, 1993) in phonology.

In the following sections, I first review this theory. Second, I sketch out a simple account of syllabification in English and show how it accounts for the linguistic facts above. Third, I show how an English syllable parser can be constructed using OT. Fourth, I show how this parser accounts for the psycholinguistic facts above. Finally, I show how this parser can be implemented computationally.

## 3. Optimality Theory

Prince & Smolensky (1993) propose that phonological generalizations should be treated in terms of constraint hierarchies rather than in terms of a sequence of rules. The basic idea is that every possible syllabification of an unsyllabified input string is generated by the function GEN and then evaluated by the function EVAL with respect to the constraint set CON. Extremely important aspects of this system are that the constraints are violable and strictly ranked. Violability means that even the best candidate may violate some constraints. Strict ranking means that the constraints are ordered such that a single violation of a higher ranked constraint is worse than any number of violations of a lower ranked constraint.

These points can be seen in the following examples. First, let's consider an abstract case. Imagine that GEN produces two candidate syllabifications for some input string. Imagine that CON consists of two constraints ranked in the order given (21) and that violations are as given.

(21)

| | /input/ | constraint 1 | constraint 2 |
|---|---|---|---|
| ☞ | candidate a | | * |
| | candidate b | *! | |

In this tableau, candidate a is judged optimal by EVAL. Only candidate b violates constraint 1 and therefore candidate a is superior. The fact that candidate a violates constraint 2 is irrelevant because constraint 2 is lower ranked than constraint 1 and the violation of constraint 1 cannot be outweighed by the violation of constraint 2, or any number of possible violations of constraint 2. This ranking of the violations is indicated in (21) with the exclamation point and shading of areas of the tableau that are then irrelevant.

Let us consider a more linguistic example. Here we adopt two constraints from the literature – FILL and ONSET. FILL requires that all segments be underlying, and thus limits epenthesis of vowels or consonants.

(22) FILL
All segments must be underlying.

The ONSET constraint requires that all syllables have onsets.

(23) ONSET
Syllables must have onsets.

Consider now how an input string like *Aggie* /ægi/ is syllabified. Syllable boundaries are marked with periods and epenthetic segments with outline.

(24)

| | /ægi/ | FILL | ONSET |
|---|---|---|---|
| ☞ a | æ.gi | | * |
| b | ℂæ.gi | *! | |
| c | æg.i | | **! |
| d | ℂæg.i | *! | * |

This tableau illustrates several points. First, note how the optimal candidate violates ONSET. Second, notice how candidate (24c) is ruled out because it violates ONSET twice, while the winning candidate only violates it once.

## 4. The basic linguistic analysis

Let us now consider how the basic facts of English can be treated in terms of OT. We will assume the constraints FILL (22) and ONSET (23). FILL captures the fact that extra segments aren't inserted all over the place; ONSET captures the essence of the MOP. We also know from the tableau in (24) that FILL is ranked above ONSET: FILL >> ONSET.

Additional constraints are also necessary as well. The PARSE constraint along with the constraints above, guarantees the correct syllabification in simple cases.

(25) PARSE
All segments must be syllabified.

The correct distribution of aspiration can be described in terms of ASPIRATION.[11]

(26) ASPIRATION
Syllable-initial voiceless stops are aspirated.

Monosyllabic words provide no additional evidence for the ranking of PARSE, FILL, ONSET, and ASPIRATION, as all these constraints are unviolated except for FILL >> ONSET as seen in (24).

(27) Ranking thus far

$$\left\{\begin{matrix}\text{PARSE}\\ \text{FILL} >> \text{ONSET}\\ \text{ASPIRATION}\end{matrix}\right\}$$

Polysyllabic words do provide evidence for the ranking of these constraints, but only when we first consider the distribution of stress. The basic analysis of stress in terms of foot structure is as follows.

[11] It is not at all clear how to extend OT to describe featural generalizations like aspiration. See, for example, Ito, Mester, & Padgett (to appear). The ASPIRATION constraint here is simply a description, as resolution of this question would take us far afield.

First, the final syllable is excluded from footing. This allows a binary foot to stress either the antepenult or penult. Classically, this was done with extrametricality (Hayes, 1982), but in OT, this is accomplished via NONFINALITY.

(28) NONFINALITY
The final syllable is not footed.

Second, a moraic trochee is built on the right edge of the word. Moraic trochees are feet that can dominate only two moras (Hayes, 1987; McCarthy & Prince, 1986). Feet are structures that dictate where stresses go and make claims about how syllables are grouped together for other phonological processes. The moraic trochee stresses the first of the two moras it dominates.

(29) FOOT-FORM (TROCHAIC)
Feet are moraic trochees.

Alignment of feet with the right edge of the word is achieved with an instance of the Generalized Alignment schema (McCarthy & Prince, 1993).

(30) ALIGN-RIGHT (WORD)
Feet are close to the right edge of the word.

The NONFINALITY constraint must be ranked above ALIGN-RIGHT to allow the optimal rightmost foot to be positioned one syllable away from the right edge (NONFINALITY >> ALIGN-RIGHT). This is shown in the following tableau with schematic representations. The two columns on the left show the correct ranking; the two columns on the right show the incorrect ranking.

(31)

| /σ σ σ σ/ | NONFINALITY | ALIGN-RIGHT | ALIGN-RIGHT | NONFINALITY |
|---|---|---|---|---|
| (σ σ)σ σ | | **! | *!* | |
| σ(σ σ)σ | | * | *! | |
| σ σ(σ σ) | *! | | | * |

Moras are supplied by vowels or (sometimes) by coda consonants. Vocalic length is universally represented with moras; it has been proposed that coda consonants are not always moraic (Hayes, 1989). The fact that vowels are universally moraic is most plausibly handled by restricting GEN. To account for the moraicity of codas, I assume that the general case is that codas are moraic.

(32) MORAICITY (preliminary version)
Coda consonants are moraic.

Extraneous moras not demanded by MORAICITY are limited by *STRUC.

(33) *STRUC
Avoid structure.

Thus MORAICITY dominates *STRUC. Our tentative ranking assumptions are given in (34).

(34) ranked
MORAICITY >> *STRUC
NONFINALITY >> ALIGN-RIGHT
ONSET >> FILL
(yet?) unranked
{ PARSE, ASPIRATION, FOOT-FORM }

Together, these moves place stress on the antepenultimate syllable unless the penult is heavy; in which case, the penult receives stress. I show how this works in the following tableaux for *agénda* and *América*.[12]

(35)

| | /agenda/ | MOR | *STRUC | NONFIN | A-R | ONSET | F-F |
|---|---|---|---|---|---|---|---|
| | (á.gen) da<br>μ μ μ | *! | | | * | | |
| | a (génda)<br>μ μ μ | *! | | | | | |
| ☞ | a (gén) da<br>μ μμ μ | | | | * | | |
| | a.gen (dá)<br>μ μμ μ | | | *! | | | * |

[12]In this and all following tableaux, I leave out PARSE and FILL, and consider no candidates that might violate those constraints.

(36)

| /america/ | MOR | *STRUC | NONFIN | A-R | ONSET | F-F |
|---|---|---|---|---|---|---|
| (áme)rica<br>μ μ μ μ | | | | **! | | |
| ☞ a(méri)ca<br>μ μ μ μ | | | | * | | |
| ame(ríca)<br>μ μ μ μ | | | *! | | | |
| ameri(cá)<br>μ μ μ μ | | | *! | | | * |

Consider now how we might treat the resyllabification facts. The basic generalization is that stressless syllables eschew onsets. This can be formalized directly as follows.

(37) NOONSET
Stressless syllables do not have onsets.

This constraint is, of course, offset by the ONSET constraint and is ranked above it. The current ranking is given in (38).

(38) ranked
MORAICITY >> *STRUC
NONFINALITY >> ALIGN-RIGHT
NOONSET >> ONSET >> FILL
(yet?) unranked
{ PARSE, ASPIRATION, FOOT-FORM }

The NOONSET constraint is new here and there are additional arguments that Universal Grammar should be augmented to include NOONSET. First, there are languages where onsetless syllables overtly and systematically reject stress, e.g. Western Aranda (Strehlow, 1942; Davis, 1985). The facts are as follows. Stress falls on even-numbered syllables from the left, except for two circumstances. First, stress never falls on the final syllable. Second, stress doesn't fall on an onsetless syllable (except to avoid final stress).

(39) túkura 'ulcer'
kútungùla 'ceremonial assistant'
wóratàra (a place name)
ergúma 'to seize'
artjánama 'to run'
utnádawàra (a place name)
káma 'to cut'
ílba 'ear'
wúma 'to hear'

These restrictions can be modeled by limiting the placement of stress with NONFINALITY and NOONSET. The fact that forms like *ílba* get initial stress shows that NONFINALITY outranks NOONSET (NONFINALITY >> NOONSET).

There is independent evidence that something like NOONSET plays a role in English. Consider the distribution of word-initial onsets in disyllabic words (from a database of 20,000 words).

(40)

| | C | V |
|---|---|---|
| iambic | 895 | 521 |
| trochaic | 3607 | 342 |

Vowel-initial iambs are a far greater percentage (37%) of the total number of iambs (1416), than vowel-initial trochees (9% of 3949). This numerical difference suggests that NOONSET plays a role.

This statistical skewing is repeated in words of all lengths. The chart in (41) shows the distribution of initial stress as a function of whether the word begins with a vowel or a consonant-vowel sequence. The column labeled "%" shows the percentage of stressless syllables for each length and segmental condition. Notice that initial stressless syllables consistently form a greater percentage of the total in the vowel-initial condition than in the consonant-initial condition.

(41)

| | CV | | | V | | |
|---|---|---|---|---|---|---|
| # of σ's | σ́ | σ̆ | % | σ́ | σ̆ | % |
| 2 | 3919 | 790 | 17 | 556 | 521 | 48 |
| 3 | 1945 | 761 | 28 | 599 | 437 | 42 |
| 4 | 346 | 663 | 66 | 116 | 400 | 78 |
| 5 | 28 | 67 | 71 | 10 | 76 | 88 |
| 6 | 0 | 10 | 100 | 0 | 10 | 100 |

Let us now consider how this system treats the kinds of items we've seen so far.[13] The following tableaux give some representative cases.

[13]For convenience, we leave out FOOT-FORM and assume all candidates make use of well-formed feet.

(42) tacky

| | /tæki/ | NONFIN | A-R | NOONS | ONSET | MOR | *STRUC |
|---|---|---|---|---|---|---|---|
| a | t$^{h}$ǽ.ki<br>\| \|<br>(µ µ) | *! | | * | | | |
| b | t$^{h}$ǽk.i<br>\|\| \|<br>(µµ µ) | *! | | | * | | * |
| c | t$^{h}$ǽk.i<br>\| \|<br>(µ µ) | *! | | | * | * | |
| d | t$^{h}$ǽ.ki<br>\| \|<br>(µ) µ | | * | *! | | | |
| ☞ e | t$^{h}$ǽk.i<br>\|\| \|<br>(µµ) µ | | * | | * | | * |
| f | t$^{h}$ǽk.i<br>\| \|<br>(µ) µ | | * | | * | *! | |
| g | t$^{h}$æ.kí<br>\| \|<br>µ (µ) | *! | | * | | | |
| h | t$^{h}$æk.í<br>\|\| \|<br>µµ (µ) | *! | | | * | | * |
| i | t$^{h}$æk.í<br>\| \|<br>µ (µ) | *! | | | * | * | |

(43) agenda

<table>
<tr><th></th><th>/æǰɛndə/</th><th>NONFIN</th><th>A-R</th><th>NOONS</th><th>ONSET</th><th>MOR</th><th>*STRUC</th></tr>
<tr><td>☞ a</td><td>əǰɛ́n.də<br>| || |<br>μ(μμ) μ</td><td></td><td>*</td><td></td><td>*</td><td></td><td>*</td></tr>
<tr><td>b</td><td>ǽǰən.də<br>| || |<br>(μ)μμ μ</td><td></td><td>**!</td><td></td><td>*</td><td></td><td>*</td></tr>
<tr><td>c</td><td>əǰən.dǽ<br>| || |<br>μ μμ (μ)</td><td>*!</td><td></td><td></td><td>*</td><td></td><td>*</td></tr>
<tr><td></td><td>əǰɛ́n.də<br>| | |<br>μ(μ μ)</td><td>*!</td><td></td><td></td><td>*</td><td>*</td><td></td></tr>
<tr><td></td><td>ǽǰən.də<br>| | |<br>(μ μ) μ</td><td></td><td>*</td><td></td><td>*</td><td>*!</td><td></td></tr>
<tr><td></td><td>əǰən.dǽ<br>| | |<br>μ μ (μ)</td><td>*!</td><td></td><td></td><td>*</td><td>*</td><td></td></tr>
</table>

Let us now consider the crucial cases. These are cases where aspiration is distributed based on the syllabic structure manipulated by NOONSET. The stress patterns of these items are not fully determined by the system we have provided. Some items have secondary stresses and some have exceptional stresses. For these items with exceptional stress, since stress is not affected by resyllabification, we consider only candidates with stress in the correct place.

(44) hotel (exceptional final stress)

<table>
<tr><th></th><th>/hotɛl/</th><th>NONF</th><th>A-R</th><th>NOO</th><th>ONS</th><th>MOR</th><th>*STR</th><th>ASPIR</th></tr>
<tr><td>☞ a</td><td>hò.tʰɛ́l<br>|\ ||<br>(μμ) (μμ)</td><td>*</td><td>*</td><td></td><td></td><td></td><td></td><td></td></tr>
<tr><td>b</td><td>hò. tɛ́l<br>|\ ||<br>(μμ) (μμ)</td><td>*</td><td>*</td><td></td><td></td><td></td><td></td><td>*!</td></tr>
<tr><td>c</td><td>hò tʰ.ɛ́l<br>|\| ||<br>(μμ) (μμ)</td><td>*</td><td>*</td><td></td><td>*!</td><td></td><td>*</td><td></td></tr>
<tr><td>d</td><td>hò t.ɛ́l<br>|\| ||<br>(μμ) (μμ)</td><td>*</td><td>*</td><td></td><td>*!</td><td></td><td>*</td><td>*</td></tr>
</table>

(45) batter

| | /bætər/ | NONF | A-R | NOO | ONS | MOR | *STR | ASPIR |
|---|---|---|---|---|---|---|---|---|
| a | bǽ.tʰər<br>\| \|<br>(μ μ) | *! | | * | | | | |
| b | bǽ.tər<br>\| \|<br>(μ μ) | *! | | * | | | | * |
| c | bǽtʰ.ər<br>\|\| \|<br>(μμ) μ | | * | | * | | * | *! |
| ☞ d | bǽt.ər<br>\|\| \|<br>(μμ) μ | | * | | * | | * | |
| e | bǽtʰ.ər<br>\| \|<br>(μ μ) | *! | | | * | * | | * |
| f | bǽt.ər<br>\| \|<br>(μ μ) | *! | | | * | * | | |

(46) attack (exceptional final stress)

| | /ətæk/ | NONF | A-R | NOO | ONS | MOR | *STR | ASPIR |
|---|---|---|---|---|---|---|---|---|
| ☞ a | ə.tʰǽk<br>\| \|\|<br>μ (μμ) | * | | | | | | |
| b | ə.tǽk<br>\| \|\|<br>μ (μμ) | * | | | | | | *! |
| c | ətʰ.ǽk<br>\|\| \|\|<br>μμ (μμ) | * | | | *! | | * | * |
| d | ət.ǽk<br>\|\| \|\|<br>μμ (μμ) | * | | | *! | | * | |
| e | ətʰ.ǽk<br>\| \|\|<br>μ (μμ) | * | | | *! | * | | * |
| f | ət.ǽk<br>\| \|\|<br>μ (μμ) | * | | | *! | * | | |

The *vanity* case is a special condition because resyllabification necessarily affects the distribution of stress and we now consider candidates with different stress patterns.

(47) vanity

| | /vænəti/ | NONF | A-R | NOO | ONS | MOR | *STR | ASPIR |
|---|---|---|---|---|---|---|---|---|
| a | vǽn.ə.tʰi<br>(μ μ) μ | | * | *! | * | * | | |
| ☞ b | vǽn.ət.i<br>(μ μ) μ | | * | | * * | * * | | |
| c | vǽn.ə.ti<br>(μ μ) μ | | * | *! | * | * | | * |
| d | vǽn.ətʰ.i<br>(μ μ) μ | | * | | * * | * * | | *! |
| e | və.ní.tʰi<br>μ (μ μ) | *! | | * * | | | | |
| f | və.nít.i<br>μ (μ μ) | *! | | * | * | * | | |
| g | və.ní.ti<br>μ (μ μ) | *! | | * * | | | | * |
| h | və.nítʰ.i<br>μ (μ μ) | *! | | * | * | * | | * |
| i | vǽn.ə.tʰi<br>(μμ) μ μ | | **! | * | * | | * | |
| j | vǽn.ət.i<br>(μμ) μμ μ | | **! | | * * | | * * | |
| k | vǽn.ə.ti<br>(μμ) μ μ | | **! | * | * | | * | * |
| l | vǽn.ətʰ.i<br>(μμ) μμ μ | | **! | | * * | | * * | * |
| m | və.nít.i<br>μ (μμ) μ | | * | *! | * | | * | |
| n | və.nítʰ.i<br>μ (μμ) μ | | * | *! | * | | * | * |

The analysis above seems to work, but does so only because the example *vánity* begins with a consonant and the NOONSET constraint eschews a stressless onsetted initial syllable: **vaníty*. The same derivation blows up with vowel-initial forms with antepenultimate stress, e.g. *ánimal*.

(48) animal

<table>
<tr><th></th><th>/ænəməl/</th><th>NONF</th><th>A-R</th><th>NOO</th><th>ONS</th><th>MOR</th><th>*STR</th><th>ASPIR</th></tr>
<tr><td>a</td><td>ǽn.ə.məl<br>|   |   |<br>(μ   μ)  μ</td><td></td><td></td><td>*!</td><td>* *</td><td>*</td><td></td><td></td></tr>
<tr><td>b</td><td>ǽn.ə.məl<br>|| |   |<br>(μμ) μ   μ</td><td></td><td></td><td>*!</td><td>* *</td><td></td><td>*</td><td></td></tr>
<tr><td>c</td><td>ǽn.əm.əl<br>|   |   |<br>(μ   μ)  μ</td><td></td><td></td><td></td><td>* * *</td><td>*!*</td><td></td><td></td></tr>
<tr><td>d</td><td>ǽn.əm.əl<br>|| |   |<br>(μμ) μ   μ</td><td></td><td></td><td></td><td>* * *</td><td>*!</td><td>*</td><td></td></tr>
<tr><td>e</td><td>æn.ə́m.əl<br>|   || |<br>μ  (μμ) μ</td><td></td><td></td><td></td><td>* * *</td><td>*!</td><td>*</td><td></td></tr>
<tr><td>✖ f</td><td>æn.ə́m.əl<br>|| || |<br>μμ (μμ) μ</td><td></td><td></td><td></td><td>* * *</td><td></td><td>* *</td><td></td></tr>
</table>

A fairly simple way of avoiding this result is to revise MORAICITY so that a following consonant is required for a consonant to be moraic.

(49) MORAICITY (revised)
preconsonantal coda consonants are moraic.

The violations of MORAICITY are then redistributed as follows.

(50) animal

<table>
<tr><th></th><th>/ænəməl/</th><th>NONF</th><th>A-R</th><th>NOO</th><th>ONS</th><th>MOR</th><th>*STR</th><th>ASPIR</th></tr>
<tr><td>a</td><td>ǽn.ə.məl<br>| | |<br>(μ μ) μ</td><td></td><td></td><td>*!</td><td>* *</td><td></td><td></td><td></td></tr>
<tr><td>b</td><td>ǽn.ə.məl<br>|| | |<br>(μμ) μ μ</td><td></td><td></td><td>*!</td><td>* *</td><td></td><td>*</td><td></td></tr>
<tr><td>☞ c</td><td>ǽn.əm.əl<br>| | |<br>(μ μ) μ</td><td></td><td></td><td></td><td>* * *</td><td></td><td></td><td></td></tr>
<tr><td>d</td><td>ǽn.əm.əl<br>|| | |<br>(μμ) μ μ</td><td></td><td></td><td></td><td>* * *</td><td></td><td>*!</td><td></td></tr>
<tr><td>e</td><td>æn.ə́m.əl<br>| || |<br>μ (μμ) μ</td><td></td><td></td><td></td><td>* * *</td><td></td><td>*!</td><td></td></tr>
<tr><td>f</td><td>æn.ə́m.əl<br>|| || |<br>μμ (μμ) μ</td><td></td><td></td><td></td><td>* * *</td><td></td><td>*!*</td><td></td></tr>
</table>

Confirmation of this approach comes from verbal and adjectival stress in English. As we saw above, stress is attracted to the penult of nouns if it is heavy; else stress falls on the antepenult. With verbs, this system shifts one syllable to the right. Stress falls on the heavy ultima of a verb; else stress falls on the penult. Presumably, this difference is a consequence of restricting the effects of NONFINALITY to nouns.

What is important in the current context, however, is that a single consonant is not sufficient to make the ultima of a verb heavy. A heavy ultima contains either a long vowel or a short vowel followed by TWO consonants.

(51)

| unsuffixed adjective | | verb | |
|---|---|---|---|
| 1 C | 2 Cs | 1 C | 2 Cs |
| stúpid | convéx | astónish | collápse |
| árid | abrúpt | consíder | tormént |
| cértain | absúrd | édit | exháust |
| cléver | bersérk | imágine | eléct |
| cómmon | corréct | intérpret | convínce |
| fóreign | delúxe | prómise | usúrp |

This difference between nouns and verbs/adjectives follows from the restriction on MORAICITY proposed above. The final consonant of words like *stupid* is nonmoraic (because of *STRUC), while the penultimate consonant of words like *convex* is moraic (because of MORAICITY).

```
(52)  stú pid    kanvéks
        |\ |       || ||
        μμ μ       μμ μμ
```

To summarize, I have proposed an OT account of aspiration and syllabification in English. This account rests largely on familiar constraints, but also invokes two novel constraints, both independently motivated. NOONSET prefers stressless syllables not to have onsets and is supported by stress in Western Aranda and statistical skewings in the English lexicon. MORAICITY is restricted to preconsonantal codas and is supported by verbal stress.

We now briefly consider how the syllabification of French can be treated.

## 5. French

French differs from English in two crucial respects. First, stress in French generally falls on the last syllable of a phrase.

(53) je ne sais pas [žənəsepá]
téléphone [telefɔ́n]
Michel [mišɛ́l]

This stress pattern can be captured by positing that the foot built in French is an iamb: a right-headed binary foot. Moreover, this iambic foot is aligned with the right edge of the phrase.

(54) FOOT-FORM (IAMBIC)
Feet are iambs.

(55) ALIGN-RIGHT (PHRASE)
Feet are at the right edge of the phrase.

French does not exhibit anything analogous to stress-based resyllabification. Instead, it is commonly assumed that medial consonant clusters are syllabified in accordance with the MOP.[14]

(56) je ne sais pas [žə.nə.se.pá]
téléphone [te.le.fɔ́n]
Michel [mi.šɛ́l]

This can be captured with the ONSET constraint.

(57) ONSET
Syllables have onsets.

[14] See, for example, Tranel (1993), but also Dell (1973).

Since French has vowel-initial words, ONSET must be ranked below FILL and PARSE.

(58) acheter [ašté]
homme [ɔ́m]
arbre [árbr]

There is no evidence that NOONSET plays any role in this system; therefore if it exists in this system, it must be placed below ONSET.

(59) ranked

$$\left\{\begin{array}{c}\text{FILL}\\ \text{PARSE}\end{array}\right\} \gg \text{ONSET} \gg \text{NOONSET}$$

unranked

$$\left\{\begin{array}{c}\text{FOOT-FORM}\\ \text{ALIGN-RIGHT}\end{array}\right\}$$

We now consider how the English and French systems might be implemented as a parser and whether a parser designed in terms of OT has any better luck in accounting for the experimental results above.

## 6. Parsing

The basic idea to be explored here is that a model of parsing that exploits ranked constraints can account for the difference between English and French illustrated above. We will adopt the most simple-minded view possible. The linguistic model above is applied to the input as soon as possible. This means that syllabification begins before the whole signal is ready for analysis and before every property of every segment is available for analysis. Thus the signal is parsed in a left-to-right fashion, and as the signal becomes clearer, the syllable parser is continually re-invoked assigning material to different syllables.

First, the basic mechanism will be that listeners parse material into well-formed syllables in a left-to-right fashion. Since they don't know what's coming before they get there, this means that there will be multiple parses available given the indeterminacy of the yet-to-be-heard material.

Second, there is no backtracking. If some particular candidate has been excluded on the basis of the parse up to some point, that candidate is not reconsidered at later points.

Third, the number of candidates is determined by the number of ways the latest increment can be adjoined to the syllable structure at that point. The parser will not analyze epenthetic elements or hypothesize unparsed segments and thus, since the number of attachment points at any point is finite, the number of candidates considered will be finite.[15]

Finally, the identification of material is a function of its serial position and the intrinsic properties being identified.

[15] I assume that resolving whether some parsed element is epenthetic or whether there are PARSE violations in the string is something left up to a later stage of processing.

Let us consider how this might work in the case of a simple CVCV string: *páta*. The parser starts off by recognizing that the signal begins with a consonant and attempts to syllabify it: CX. Specific properties of this consonant will trickle in gradually as they are determined, but in this case since the precise identity of this consonant makes no difference these factors won't affect anything.

The parser next determines that the following segment is a vowel and attempts to parse it as well: pVX. This process continues until the signal ends. The unanalyzed signal is denoted with 'X'.

(60) X -> CX -> pX -> pVX -> páX -> páCX -> pátX -> pátV -> páta

Note that this process makes the assumption that the contrast between vowels and consonants is available to the parser before other contrasts. This particular distinction is not crucial. What is crucial to the system developed here is that the distinction between stressed and stressless vowels in English emerges after other distinctions, like the distinction between consonants and vowels.[16]

Each stage of the process involves applying GEN to generate a new set of candidates. The constraint hierarchy CON evaluates that set of candidates. Let us go through the schematic example above with a simplified constraint hierarchy.

(61) Simplified hierarchy for English
PARSE >> NOONSET >> ONSET

We can compare the parse of a similar string in French. As argued above, the constraint NOONSET is absent in French or ranked below ONSET. (Ranking NOONSET below ONSET renders it completely redundant.)

(62) Simplified hierarchy for French
PARSE >> ONSET >> NOONSET

The parser first recognizes that the string begins with a consonant: X -> CX. In this case, there are only two possibilities, Either the consonant is parsed or not. Since PARSE is the highest constraint and no following material need be considered to evaluate violations of parse, the first consonant ends up being syllabified. Constraints that cannot be evaluated because of insufficient context are shaded and violations not marked.

(63) X -> CX

| /CX/ | PARSE | NOONSET | ONSET |
|---|---|---|---|
| CX<br>\\<br>σ | | | |
| CX | *! | | |

[16] We return to this assumption below.

The consonant is eventually identified and its identity does not affect the parse. Next, the following vowel is considered. In this case, the affiliations of the consonant and vowel do not depend on what follows and therefore a determinate parse is reached.

(64) pX -> pVX

<table>
<tr><th></th><th>/pVX/<br>\<br>σ</th><th>PARSE</th><th>NOONSET</th><th>ONSET</th></tr>
<tr><td>☞ a</td><td>pVX<br>\|<br>σ</td><td></td><td></td><td></td></tr>
<tr><td>b</td><td>pVX<br>\<br>σ</td><td>*!</td><td></td><td></td></tr>
</table>

Violations of NOONSET cannot be determined because the stressedness of this vowel isn't yet known. By transitivity, since ONSET is ranked below NOONSET, its violations are also irrelevant. However, PARSE provides for a unique best candidate since there are only two candidates and one of them violates PARSE. Since candidates (64b) is excluded in this tableau, it is not considered in any further tableaux.

When the vowel is identified in (65), there is only one candidate and nothing changes.

(65) pVX -> páX

<table>
<tr><th></th><th>/páX/<br>\|<br>σ</th><th>PARSE</th><th>NOONSET</th><th>ONSET</th></tr>
<tr><td>☞ a</td><td>páX<br>\|<br>σ</td><td></td><td></td><td></td></tr>
</table>

Tableau (66) shows the next stage where the parser determines that there is a following consonant. Notice that violations of neither ONSET or NOONSET can be determined since the identity of the following material is as yet unclear. This results in two optimal candidates.

(66) páX -> páCX

<table>
<tr><th></th><th>/páCX/<br>\|<br>σ</th><th>Parse</th><th>NoOnset</th><th>Onset</th></tr>
<tr><td>☞ a</td><td>páCX<br>\|\<br>σ σ</td><td></td><td></td><td></td></tr>
<tr><td>☞ b</td><td>páCX<br>\|/<br>σ</td><td></td><td></td><td></td></tr>
<tr><td>c</td><td>páCX<br>\|<br>σ</td><td>*!</td><td></td><td></td></tr>
</table>

In the next step, the identity of the consonant becomes known, but this makes no difference for the constraints at issue and we therefore leave out this step.

Next a following vowel is extracted from the signal. Notice that there are now two possible input forms (because the preceding tableau provided two tied outputs). Again, violations of NoOnset cannot be assessed as the stressedness of this second vowel is not yet available. By transitivity, violations of Onset are also irrelevant.

(67) pátX -> pátV

<table>
<tr><th></th><th>/pátV/ or /pátV/<br>\|\ \|/<br>σ σ σ</th><th>Parse</th><th>NoOnset</th><th>Onset</th></tr>
<tr><td>a</td><td>pátV<br>\|\<br>σ σ</td><td>*!</td><td></td><td></td></tr>
<tr><td>☞ b</td><td>pátV<br>\|\|<br>σ σ</td><td></td><td></td><td></td></tr>
<tr><td>☞ c</td><td>pátV<br>\|/\<br>σ σ</td><td></td><td></td><td>*</td></tr>
<tr><td>d</td><td>pátV<br>\|/<br>σ</td><td>*!</td><td></td><td></td></tr>
</table>

In the final stage of the derivation, the identity of the final vowel becomes available and allows violations of NoOnset to be determined.

(68) pátV -> páta

<table>
<tr><th></th><th>/páta/ or /páta/<br>\|\|     \|/\<br>σ σ     σ σ</th><th>PARSE</th><th>NOONSET</th><th>ONSET</th></tr>
<tr><td>a</td><td>páta<br>\|\|<br>σ σ</td><td></td><td>*!</td><td></td></tr>
<tr><td>☞ b</td><td>páta<br>\|/\<br>σ σ</td><td></td><td></td><td>*</td></tr>
</table>

It is thus only with the full identification of the postconsonantal vocalic segment – specifically whether it is stressed or not – that the parse can resolve to a single candidate in English.

Consider now how this process might work in French. As noted above, the crucial difference between French and English is that NOONSET is either absent or ranked below ONSET in French. What this means is that the syllabification of CVCV strings can resolve to a single candidate sooner in French. The relevant point of the parse is /pátV/. Neither system can resolve to a single parse with /pátX/ and both resolve with /páta/.

(69) pátX -> pátV

<table>
<tr><th></th><th>/pátV/ or /pátV/<br>\|\     \|/<br>σ σ     σ</th><th>PARSE</th><th>ONSET</th><th>NOONSET</th></tr>
<tr><td>a</td><td>pátV<br>\|\<br>σ σ</td><td>*!</td><td></td><td></td></tr>
<tr><td>☞ b</td><td>pátV<br>\|\|<br>σ σ</td><td></td><td></td><td></td></tr>
<tr><td>c</td><td>pátV<br>\|/\<br>σ σ</td><td></td><td>*!</td><td></td></tr>
<tr><td>d</td><td>pátV<br>\|/<br>σ</td><td>*!</td><td></td><td></td></tr>
</table>

As in English, violations of NOONSET cannot be determined, but because it is ranked below ONSET, a single parse can be found.

What this means is that because of the differing phonological systems of English and French – (61) vs. (62) – and the relative lateness with which crucial information for determining the relevance of NOONSET is available, the languages differ in the rate with which a successful parse is reached. In English, a successful parse can be reached only when the stress of the second vowel is available; in French, the stress of the second vowel is irrelevant.

What this means is that if syllabic information is used during the fragment monitoring task, a determinate syllabification is reached later in English than in French. Following much work in this domain, I assume that there are two routes subjects take in identifying target strings: a segmental/phonemic route and a syllable-based route. In French, presumably the syllabification route succeeds before the segmental route; in English, presumably, the syllabification route does not resolve soon enough. Hence there is no effect of syllabification.

The same point can be made with the other contrast in English: *clímàx* vs. *clímate*. The relevant tableaux is given below. At the point where the segments have been identified and the C/V status of the following segment is clear, the syllabification of either form is still indeterminate.

(70) kláymX -> kláymV

<table>
<tr><th></th><th>/kláymV/ or /kláymV/<br>\\|/\       \\|//<br>σ σ       σ</th><th>PARSE</th><th>NOONSET</th><th>ONSET</th></tr>
<tr><td>a</td><td>kláymV<br>\\|\<br>σ σ</td><td>*!</td><td></td><td></td></tr>
<tr><td>☞ b</td><td>kláymV<br>\\|/|<br>σ σ</td><td></td><td></td><td></td></tr>
<tr><td>☞ c</td><td>kláymV<br>\\|/\<br>σ σ</td><td></td><td></td><td>*</td></tr>
<tr><td>d</td><td>kláymV<br>\\|/<br>σ</td><td>*!</td><td></td><td></td></tr>
</table>

The basic machinery exploited to account for the difference between English and French is twofold. There is a phonological distinction and a processing distinction. The phonological difference has been argued for in §4.

The processing distinction is as follows. I have proposed that material is syllabified as it becomes clear and have proposed that the distinction between consonants and vowels is recovered relatively early and that whether a vowel is stressed or not is recovered later. Do we have any reason to believe this is so?

Yes.

First, it is surely uncontroversial that different aspects of the speech stream should exhibit different degrees of salience and be extractable at different rates.

Second, a priori, the difference between consonants and vowels would be expected to be a salient one. Vowels exhibit a full formant structure while consonants either do not, or exhibit a formant structure that is reduced in some fashion (nasals and liquids). First, these distinctions are readily apparent over the duration of segments in that the full formant structure of a vowel persists over the duration of the vowel. Other distinctions like place of articulation or vowel height are much more fleeting. Second, these distinctions are not relative, while other distinctions – like vowel height or stress – are.

Second, there are, in fact, minimal pairs in English where the stress of a vowel is only evident because of the cues resulting from ASPIRATION. For example:

| | | | | |
|---|---|---|---|---|
| (71) | mánatèe | [mǽnət<sup>h</sup>ì] | vánity | [vǽnəɾi] |
| | vétò | [vít<sup>h</sup>ò] | mótto | [máɾo] |

An additional source of evidence for the salience of the CV distinction is that there are other systems that derive this result. For example, Sejnowsky & Rosenberg (1986) propose a connectionist architecture system – NETtalk – that learns the grapheme-to-phoneme conversion rules of English and produces synthetic output. They argue that "...few errors in a well-trained network were confusions between vowels and consonants: most confusions were between phonemes that were very similar..."(p.667).

Obviously more study is required here, but the basic idea is almost definitionally true in some respect: certain qualities of the signal will emerge before others as a consequence of their salience in the signal and the limits of the human perceptual apparatus.

## 7. The implementation

The model described above is implemented in Perl (Wall & Schwartz, 1991) and makes use of several innovations that result in a relatively efficient resolution of the candidate set.[17] There are three main ones.

(72) a. finite candidate set
b. serial constraint satisfaction
c. local coding

In this section, I first outline the program and then show how each of the above works to make EVAL more efficient.

The program works as follows. First, the "language" module prompts the user to provide a language and the "getword" module prompts for an input word. The language response tells the program whether to adopt the English (61) or French (62) constraint rankings. The "addcv" module then analyzes the input and the "submitword" module submits it to the parser in the fashion described above. The "mhgen" subroutine generates the candidate sets each time a new segment is added to the input.

[17] See Black, (1993), Ellison (1995), Kitahara (1994), and Tesar (1995) for previous work on implementing OT.

The candidate sets are coded in a novel fashion. Recall, for example, the sample input *páta* in (60) above, repeated as (73) below.

(73) X -> CX -> pX -> pVX -> páX -> páCX -> pátX -> pátV -> páta

Each segment can be syllabified as an onset ("o"), coda ("c"), nucleus ("n"), or be unsyllabified ("u"). Each segment of the input is coded for these four options and then options are gradually removed by the constraint system. The analysis of (73) would proceed as in (74) below.

(74)

| | | | | | | | | | | | | | | | |
|---|---|---|---|---|---|---|---|---|---|---|---|---|---|---|---|
| oncu | $\xrightarrow{a}$ | o | $\xrightarrow{b}$ | o | $\xrightarrow{c}$ | n/a | $\xrightarrow{d}$ | o oncu | $\xrightarrow{e}$ | o n | $\xrightarrow{f}$ | o n | $\xrightarrow{g}$ | n/a | $\xrightarrow{h}$ |
| C | | C | | p | | | | p V | | p V | | p á | | | |

| | | | | | | | | | | |
|---|---|---|---|---|---|---|---|---|---|---|
| $\xrightarrow{h}$ | o n oncu | $\xrightarrow{i}$ | o n oc | $\xrightarrow{j}$ | o n oc | $\xrightarrow{k}$ | n/a | $\xrightarrow{l}$ | o n oc oncu | $\xrightarrow{m}$ |
| | p á C | | p á C | | p á t | | | | p á t V | |

| | | | | | |
|---|---|---|---|---|---|
| $\xrightarrow{m}$ | o n oc n | $\xrightarrow{n}$ | o n oc n | $\xrightarrow{o}$ | o n c n |
| | p á t V | | p á t a | | p á t a |

The "doconstraints" module eliminates options on each pass. It is important to note that each segment resolves to a single candidate parse by the time it is the penultimate segment in the string under consideration.

Let us now consider the three salient properties of this model.

First, the candidate set is always finite. Recall that this is because the parser does not hypothesize epenthetic segments or unparsed segments. This is clearly a contributor to resolving an optimal parse in finite time.

Second, this model exhibits serial constraint satisfaction. Given that definitive parses are reached for any segment by the time it is the penultimate segment of the string, this means that the parser must consider a maximum of 8 candidate parses for any input (assuming that all four parses are still available for the final two segments – which is never the case in the current implementation). The maximum number of passes is a function of the number of segments in the word. For example, a word with four segments will undergo 8 passes through the constraint system. Assuming (falsely) that the 8 maximum candidates per pass were disjoint, this means that the parser will consider a maximum of 64 candidates, or $(2n)^2$, where $n$ = segments. In the example above, coding the four options for each segment in a single list of candidates would entail $4^4$ = 256 candidates to consider. Thus serial constraint satisfaction results in a significant reduction in the number of candidates considered.[18]

Finally, candidates are coded locally. This "local coding" of the constraint set effectively means that even fewer candidates are considered in toto. Even if all options were considered at once, the system would be coping with at most 16 options.

---

[18]This seems similar at least in spirit to what Tesar (1995) refers to as "dynamic programming".

(75) oncu oncu oncu oncu
p á t a

In fact, at any one point, the maximum number of competing candidates considered is only 6 and the sum total across all the stages of the derivation is only 26.

Thus, the implementation proposed drastically reduces the number of candidates that must be considered. It does so by i) generating only a finite number of candidate parses, ii) eliminating candidates in a serial fashion, and ii) coding candidate parses locally.

## 8. Conclusion

I have proposed a specific linguistic analysis of aspiration and syllabification in English. This analysis was built on familiar constraints from the OT literature, except for two. First, I proposed the NOONSET constraint which requires that stressless syllables be onsetless. This constraint received additional motivation from Western Aranda stress and distributional regularities of the English lexicon. Second, I proposed the MORAICITY constraint, which requires that preconsonantal codas be moraic.[19] This constraint was supported by additional facts concerning verbal stress.

This analysis was then extended to provide an account of some problematic psycholinguistic results. Prima facie, these fragment monitoring results suggest that the syllable is made use of in the processing of French, but not in the on-line processing of English. This processing difference emerges naturally from the phonological differences between French and English syllabification and from the time course of segmental identification.

This is an important result because i) it supports the emerging Optimality Theory, and ii) it shows that results in linguistics and psycholinguistics can be put together to make progress.

Finally, the model was implemented in a computer program. The implementation shows that the parser is feasible and demonstrates the efficiency of i) a finite candidate set, ii) serial constraint satisfaction, and iii) local coding of candidates.

## 9. Appendix A: Code

The model described above has been implemented using Perl (Practical Extraction and Report Language). This language was used because of its excellent pattern-matching resources and because it is widely available as shareware/freeware for different architectures (at least UNIX, DOS, and Macintosh). The code is given below.[20]

```
#!/usr/local/bin/perl

###################sylpars##################

#This is a fragment of a constraint-based syllable
```

[19]See Fountain (1994) for another OT-based treatment of moraic codas.
[20]This can also be obtained via WWW at http://aruba.ccit.arizona.edu/~hammond.

```
#parser written by Mike Hammond in Perl. This code
#is described in "Syllable parsing in English and
#French". If you know nothing of Perl, this code can
#be run in most UNIX systems by making this file
#executable ('chmod u+x sylpars') and then invoking
#it directly. The program can take any alphabetic string
#as input (though it doesn't understand anything beyond
#the difference between consonants and vowels). Stress
#is indicated with an apostrophe preceding the
#relevant vowel. The program will run with either
#French or English ranking on any input. For example,
#to run 'hat' with English ranking, type:

#sylpars "e" "h'at"

#Output can be saved by redirection to a file,
#e.g.

#sylparse "e" "h'at" > myfile

#The program can be invoked without specifying the
#langauge or input on the command line, but then output
#cannot be saved as easily.

#Mike Hammond
#Linguistics
#University of Arizona
#Tucson, AZ 85721

#hammond@aruba.ccit.arizona.edu

##################################beginning of program

#debug flag (for annoyingly verbose output)

#$mhdebug = 1;

#################################beginning of main

#sets the language

&language;

#this gets the word

&getword;

#make the cv template....

&addcv;

#submits the word to the constraints

&submitword;
```

```
print "#" x 20;
print "\n\nword: $theword";
print "language: $language\n";
print "All done!\n\n";


#####################################end of main

#sets the flag controlling the constraint ranking

sub language {
   if ($ARGV[0] =~ /^$/) {
     print "\nEnglish (e: default) or French (f)?:";
     $language = <STDIN>;
   }
   else {
     $language = $ARGV[0];
   }
   if ($language =~ /^$/) { $language = "e"; }
   if ($language =~ /^[eE]$/) { $language = 'English'; }
   elsif ($language =~ /^[fF]$/) { $language = 'French'; }
   else { die "You can only enter \"e\" or \"f\".\n"; }
}

#this gets the word to be parsed

sub getword {
  if ($ARGV[1] ne "") {
     $theword = $ARGV[1] . "\n";
  }
  else {
    print "\nEnter word: ";
    $theword = <STDIN>;
  }

  #this splits the characters up and puts them in an array

  $size = length($theword) - 1;
  if (defined($mhdebug)) {
     print "number of characters: $size\n";
     }
  for ($i = 0; $i < $size; $i++) {
     $mhinput[$i] = substr($theword, $i, 1);
     }

  #this checks that letters are letters

  for ($i = 0; $i <= $#mhinput; $i++) {
     $mhinput[$i] =~ /[a-zA-Z\']/ || die "Illegal characters!\n";
     }

  #this copies the stress mark into the same
  #array as the following vowel
```

```
  for ($i = 0; $i <= $#mhinput; $i++) {
     if ($mhinput[$i-1] =~ /\'/) {
        $mhinput[$i] =~ s/([aeiouyAEIOUY])/\'$1/;
        }
  }

  #this removes stressmarks alone

  foreach $letter (@mhinput) {
      if ($letter !~ /\'$/) {
         push(@tempinput, $letter);
      }
  }
  @mhinput = @tempinput;
}

#creates the cv array

sub addcv {
     foreach $let (@mhinput) {
       if ($let =~ /[aeiouyAEIOUY]/) {
           push (@mhcv, 'V');
       }
       else {
           push (@mhcv, 'C');
       }
     }
  if (defined($mhdebug)) {
     print "CV skeleton:";
     print join (" ", @mhcv);
     print "\n\n";
     }
}

#simulates the left to right identification of
#segments and the gradual identification of
#individual segments

sub submitword {
    $pass = 1;
    for ($i = 0; $i <= $#mhinput; $i++) {

       #adds unidentified CVs to the input string

       push(@submission, ($mhcv[$i]));
       push(@structure, '-');
       if (defined($mhdebug)) {
         print "submitting...\n";
       }

       print "parse #$pass\n";
       $writething = join(" ", @submission);
       $writestruc = join(" ", @structure);
```

```
        print "$writestruc\n";
        print "$writething\n\n";

        #generates candidate syllabifications

        &mhgen;

        #eliminates candidates via eval

        &doconstraints;

        #replaces CVs with real consonants and vowels

        splice(@submission, $#submission, 1, $mhinput[$i]);

        if (defined($mhdebug)) {
          print "submitting...\n";
        }

        print "parse #$pass\n";
        $writething = join(" ", @submission);
        $writestruc = join(" ", @structure);
        print "$writestruc\n";
        print "$writething\n\n";

        #eliminates more candidates with eval

        &doconstraints;
    }
}

#this figures out the optimal syllabification
#parse >> noonset >> onset in English
#parse >> onset >> noonset in French

sub doconstraints {
     print "pass #$pass\n";
     $pass++;
     print "input candidate(s):\n";
     $writething = join(" ", @submission);
     $writestruc = join(" ", @structure);
     print "$writestruc\n";
     print "$writething\n\n";

  #the basic constraint set is in dobasictypes and the
  #different rankings in the two languages in ordering of
  #the different specific constraints following

  print "output candidate(s):\n";
  &dobasictypes;
  if ($language eq "English") {
     &doparse;
     &donoonset;
     if ($skipflag != 1) {
```

```
        &doonset;
      }
    }
    else {
      &doparse;
      &doonset;
      &donoonset;
    }

    $skipflag = 0;

    if (defined($mhdebug)) {
        print "outputting...\n";
    }
    $writething = join(" ", @submission);
    $writestruc = join(" ", @structure);
    print "$writestruc\n";
    print "$writething\n\n";
}

#the following creates the candidate syllabifications for
#different segments. Coding for syllabic position:
# o: onset, n: nucleus, c: coda,
# -: unanalyzed, u: unsyllabified

sub mhgen {
  if (defined($mhdebug)) {
      print "generating candidate set...\n";
  }
  splice(@structure, $#structure, 1, 'oncu');
}

#the following handles a bunch of generalizations
#not treated by the explicit constraints and entailed
#by the local coding above.

sub dobasictypes {
  if (defined($mhdebug)) {
       print "checking basic stuff...\n";
  }

   #vowels can't be onsets or codas

   if ($submission[$#submission] =~ /V/) {
      $structure[$#structure] =~ s/[oc]//g;
   }

   #consonants can't be nuclei

   if ($submission[$#submission] =~ /C/) {
      $structure[$#structure] =~ s/n//;
   }

   #word-initial consonants can't be codas
```

```
    if ($#structure == 0) {
        $structure[0] =~ s/(.)c/$1/;
    }

    #word-final consonants can't be onsets

    if ($#structure == $#mhinput) {
        $structure[$#mhinput] =~ s/o(.)/$1/;
    }

    #preconsonantal consonants can't be onsets

    if ($submission[$#submission] =~ /C/ && $#submission > 0) {
        $structure[$#structure-1] =~ s/o(.)/$1/;
    }

    #postonset consonants can't be codas

    if ($structure[$#structure-1] =~ /o/) {
        $structure[$#structure] =~ s/(.)c/$1/;
    }

}

#PARSE eliminates 'u' as a candidate if other parses
#are available

sub doparse {
  if (defined($mhdebug)) {
     print "checking PARSE...\n";
  }
  $structure[$#structure] =~ s/(.)u/$1/;
}

#NOONSET eliminates 'o' as an option for the preceding
#segment if the current segment is a stressless vowel

sub donoonset {
  if (defined($mhdebug)) {
    print "checking NOONSET...\n";
  }
  if ($submission[$#submission] =~ /^[aeiouyAEIOUY]$/) {
     if ($#submission > 0) {
        $structure[$#structure-1] =~ s/o(.)/$1/;
     }
  }
  if ($submission[$#submission] =~ /V/) {
    $skipflag = 1;
  }
}

#ONSET eliminates 'c' as an option if the current segment
#is a vowel
```

```
sub doonset {
  if (defined($mhdebug)) {
    print "checking ONSET...\n";
  }
  if ($submission[$#submission] =~ /V/ ||
                       $submission[$#submission] =~ /[aeiouyAEIOUY]/) {
      if ($#submission > 0) {
        $structure[$#structure-1] =~ s/(.)c/$1/;
      }
  }
}
```

## 10. Appendix B: sample output

Here we give sample output for the program with English and French constraint rankings with the following hypothetical inputs: /apa/ and /apá/. The critical stages where the French parse is fixed but the English parse is not are given in bold face.

*English /apa/*

```
English (e: default) or French (f)?:

Enter word: apa
parse #1
-
V

pass #1
input candidate(s):
oncu
V

output candidate(s):
n
V

parse #2
n
a

pass #2
input candidate(s):
n
a

output candidate(s):
n
a

parse #3
n -
a C
```

```
pass #3
input candidate(s):
n oncu
a C

output candidate(s):
n oc
a C

parse #4
n oc
a p

pass #4
input candidate(s):
n oc
a p

output candidate(s):
n oc
a p

parse #5
n oc -
a p V

pass #5
input candidate(s):
n oc oncu
a p V

output candidate(s):
n oc n
a p V

parse #6
n oc n
a p a

pass #6
input candidate(s):
n oc n
a p a

output candidate(s):
n c n
a p a

####################

word: apa
language: English
All done!
```

*English: /apá/*

```
English (e: default) or French (f)?:

Enter word: ap'a
parse #1
-
V

pass #1
input candidate(s):
oncu
V

output candidate(s):
n
V

parse #2
n
a

pass #2
input candidate(s):
n
a

output candidate(s):
n
a

parse #3
n -
a C

pass #3
input candidate(s):
n oncu
a C

output candidate(s):
n oc
a C

parse #4
n oc
a p

pass #4
input candidate(s):
n oc
a p
```

```
output candidate(s):
n oc
a p

parse #5
n oc -
a p V

pass #5
input candidate(s):
n oc oncu
a p V

output candidate(s):
```
**n oc n**
**a p V**

```
parse #6
n oc n
a p 'a

pass #6
input candidate(s):
n oc n
a p 'a

output candidate(s):
n o n
a p 'a

####################

word: ap'a
language: English
All done!
```

*French: /apa/*

```
English (e: default) or French (f)?:f

Enter word: apa
parse #1
-
V

pass #1
input candidate(s):
oncu
V

output candidate(s):
n
```

V

parse #2
n
a

pass #2
input candidate(s):
n
a

output candidate(s):
n
a

parse #3
n –
a C

pass #3
input candidate(s):
n oncu
a C

output candidate(s):
n oc
a C

parse #4
n oc
a p

pass #4
input candidate(s):
n oc
a p

output candidate(s):
n oc
a p

parse #5
n oc –
a p V

pass #5
input candidate(s):
n oc oncu
a p V

output candidate(s):
**n o n**
**a p V**

```
parse #6
n o n
a p a

pass #6
input candidate(s):
n o n
a p a

output candidate(s):
n o n
a p a

####################

word: apa
language: French
All done!
```

*French: /apá/*

```
English (e: default) or French (f)?:f

Enter word: ap'a
parse #1
-
V

pass #1
input candidate(s):
oncu
V

output candidate(s):
n
V

parse #2
n
a

pass #2
input candidate(s):
n
a

output candidate(s):
n
a

parse #3
n -
a C
```

pass #3
input candidate(s):
n oncu
a C

output candidate(s):
n oc
a C

parse #4
n oc
a p

pass #4
input candidate(s):
n oc
a p

output candidate(s):
n oc
a p

parse #5
n oc –
a p V

pass #5
input candidate(s):
n oc oncu
a p V

output candidate(s):
**n o n**
**a p V**

parse #6
n o n
a p 'a

pass #6
input candidate(s):
n o n
a p 'a

output candidate(s):
n o n
a p 'a

###################

word: ap'a
language: French
All done!

Michael Hammond
Department of Linguistics
University of Arizona
Tucson, AZ 85721

email: hammond@aruba.ccit.arizona.edu